\documentclass[12pt]{article}
\usepackage[utf8]{inputenc}
\usepackage[T1]{fontenc}
\usepackage{amssymb}
\usepackage{amsmath}
\usepackage{stfloats}
\usepackage [autostyle, english = american]{csquotes}
\MakeOuterQuote{"}
\usepackage{subcaption}
\usepackage{textcomp}
\usepackage{setspace}
\usepackage{blindtext}
\usepackage{graphicx}
\usepackage{lettrine}
\usepackage{lipsum}
\usepackage{multicol}
\usepackage{multirow}
\usepackage{textcomp}
\usepackage{layouts}
\usepackage[font=small,skip=0pt]{caption}
\usepackage{longtable}
\usepackage{color}
\usepackage{mathtools}
\usepackage{booktabs}

\usepackage{array}
\newcolumntype{L}[1]{>{\raggedright\let\newline\\\arraybackslash\hspace{0pt}}m{#1}}
\newcolumntype{C}[1]{>{\centering\let\newline\\\arraybackslash\hspace{0pt}}m{#1}}
\newcolumntype{R}[1]{>{\raggedleft\let\newline\\\arraybackslash\hspace{0pt}}m{#1}}
\usepackage[linesnumbered]{algorithm2e}
\usepackage{bbm}
\usepackage{duckuments}
\usepackage[hidelinks]{hyperref}
\usepackage[left=2cm, right=3cm, top=2cm, bottom=2.5cm]{geometry}
\usepackage{authblk}
\usepackage{epstopdf}
\usepackage{etoolbox}
\makeatletter
\def\@seccntformat#1{\@ifundefined{#1@cntformat}%
   {\csname the#1\endcsname.\hskip0.5em}    
   {\csname #1@cntformat\endcsname}
}
\appto{\appendix}{%
    \renewcommand{\appendixname}{Appendix}
    \newcommand{\section@cntformat}{\appendixname~\Alph{section}.\hskip0.5em}
    \newcommand{\subsection@cntformat}{\Alph{section}.\arabic{subsection}.\hskip0.5em}}
\makeatother

\begin{document}

\title{Reinforcement learning adaptive fuzzy controller for lighting systems: application to aircraft cabin}

\author[1]{Kritika Vashishtha}
\author[1]{Anas Saad}
\author[1,*]{Reza Faieghi}
\author[1]{Fengfeng Xi}
\affil[1]{\small{Department of Aerospace Engineering, Toronto Metropolitan University, 350 Victoria St., M5B2K3, Toronto, Ontario, Canada}}
\affil[*]{Corresponding author: reza.faieghi@torontomu.ca}
\date{}

\maketitle

\begin{abstract}
The lighting requirements are subjective and one light setting cannot work for all. 
However, there is little work on developing smart lighting algorithms that can adapt to user preferences.
To address this gap, this paper uses fuzzy logic and reinforcement learning to develop an adaptive lighting algorithm.
In particular, we develop a baseline fuzzy inference system (FIS) using the domain knowledge.
We use the existing literature to create a FIS that generates lighting setting recommendations based on environmental conditions i.e. daily glare index, and user information including age, activity, and chronotype.
Through a feedback mechanism, the user interacts with the algorithm, correcting the algorithm output to their preferences.
We interpret these corrections as rewards to a Q-learning agent, which tunes the FIS parameters online to match the user preferences.
We implement the algorithm in an aircraft cabin mockup and conduct an extensive user study to evaluate the effectiveness of the algorithm and understand its learning behavior.
Our implementation results demonstrate that the developed algorithm possesses the capability to learn user preferences while successfully adapting to a wide range of environmental conditions and user characteristics.
and can deal with a diverse spectrum of environmental conditions and user characteristics.
This underscores its viability as a potent solution for intelligent light management, featuring advanced learning capabilities.


\end{abstract}


\section{Introduction}\label{se:Intro}
Interior lighting plays a crucial role in enhancing the comfort and well-being of aircraft passengers. 
Diverse lighting scenarios create a variety of atmospheres and experiences for passengers, and can significantly impact passengers' mood, perception, alertness, and circadian rhythm. This can range from creating a sense of relaxation to providing stimulation or entertainment.
Additionally, lighting can assist passengers in adapting to different time zones and minimizing the effects of jet lag \cite{vink2012possibilities, vink2016aircraft}.

With the advancement of artificial intelligence and control systems, intelligent lighting systems are becoming increasingly popular.
Previous research in this area includes the development of a smart lighting system to elevate visual comfort while minimizing energy consumption \cite{ciabattoni2013smart}.
The system maintains a desired light level where needed while minimizing it where not required.
This concept is further explored in \cite{sikder2018iot}, using the Internet of Things (IoT) to create an autonomous and more efficient lighting management system in smart cities.
Another notable work is \cite{xu2019design}, which details the design, implementation, and deployment of a smart emergency light system for buildings.
The system’s key advantage is its integration with existing building facilities (i.e., emergency lights), and it has been successfully implemented in smart buildings.

While developing similar systems for aircraft interiors holds great promise for elevating the flight experience, progress in this area has been limited.
The papers that we were able to find have primarily focused on the architecture of intelligent lighting systems with little or no emphasis on developing intelligent light management algorithms \cite{zeng2022design, wan2022civil, gu2006intelligent}.
Our objective in this paper is to present an intelligent algorithm for light management in aircraft cabins.
In particular, we aim to develop an algorithm that can automatically adjust the intensity of aircraft interior lights based on the perception of the environment, e.g., outdoor lighting and passengers' activities.
Further, the algorithm should be able to learn the passengers' preferences and adapt light settings to their liking.

A particular challenge in developing such an algorithm is that lighting preferences are highly individualistic and influenced by a multitude of factors.
As pointed out in \cite{ochoa2012considerations}, factors such as age, gender, and even the type of activity being undertaken can significantly impact an individual's lighting needs.
For instance, the light intensity required for reading a book can greatly differ from the light intensity preferred while watching a movie or enjoying a meal.
Further, visual comfort and glare sensations are subjective feelings that cannot be accurately captured by a one-size-fits-all setting.

In this intricate landscape where a comprehensive model is difficult to find, we propose to use a rule-based algorithm, namely a fuzzy inference system (FIS). 
The strength of a FIS lies in its ability to build a controller using intuitive natural rules. 
This becomes particularly advantageous in controlling processes that are difficult to model, such as the nuanced interplay of environmental conditions and personal preferences in defining optimal lighting conditions.
In fact, FIS has been successfully implemented in related contexts such as IoT-based traffic light control \cite{chiesa2020fuzzy}, and also energy-saving strategies in smart LED lighting systems that account for lighting comfort and daylight \cite{liu2016fuzzy}.
In addition, our team has had success in using FIS to automatically adjust the transparency of electrochromic windows and control glare in aircraft cabins to maintain a comfortable visual environment amidst varying glare conditions \cite{jasper}.

As we intend to develop an adaptive algorithm that can learn and respond to the user's preferences, we need to enhance FIS with an adaptive mechanism that can interact with the user, interpret the interaction, and take action accordingly. 
A standard method of choice to achieve this is reinforcement learning (RL) \cite{sutton2018reinforcement}.
The combination of RL and FIS is an active area of research with a wide range of applications; see for example \cite{chen2023reinforcement, li2022using, zhu2022fuzzy, guo2022energy, ghaderi2022power}.
The choice of the RL algorithm depends on the inherent characteristics of the problem domain. 
As will be clarified in our developments, for the intelligent lighting management system, the problem domain is relatively small, discrete, and model-free.
One RL algorithm that is proven effective for such problems is Q-learning 
 \cite{sutton2018reinforcement}.
Therefore, a Q-learning FIS; hereafter referred to as QFIS, will be our method of choice to develop the intended adaptive light management algorithm.
QFIS has been explored in numerous studies and has been successfully implemented for various applications; see for example \cite{desouky2011self, desouky2011q, al2014investigation, al2014two, bo2022q, yang2020q, iskandar2021q}.
The idea behind QFIS is to build a self-learning FIS whose parameters can be tuned online using Q-learning.

In this study, we first develop a FIS for aircraft cabin light management.
We develop the FIS using domain knowledge and consider inputs such as age, chronotype, and user activity type.
Next, we make the FIS adaptive by designing a Q-learning algorithm that adjusts the FIS parameters online during interactions with the user.
We present an extensive user study in an aircraft cabin to evaluate the effectiveness of the algorithm.
In our developments, we adopt the standard formulation of QFIS given in \cite{desouky2011q}; however, this requires careful attention in determining the inputs, membership functions (MFs), FIS rules, the representation of problem domain using Q-tables, the choice of a reward function, and the learning rate, all of which are detailed in subsequent sections.

The major contribution of this work is to develop an intelligent lighting management algorithm for aircraft cabins.
To our knowledge, this algorithm is the first to automatically change lighting settings based on user activities and preferences.
While our focus is on aircraft interiors, the developments here can be extended to different lighting where an intelligent light management system is desired, e.g., smart buildings.
\section{Algorithm Overview}\label{se:overview}
\begin{figure}[h]
    \centering
    \includegraphics[trim={0cm 6cm 11cm 0cm}, clip, width=0.8\textwidth]{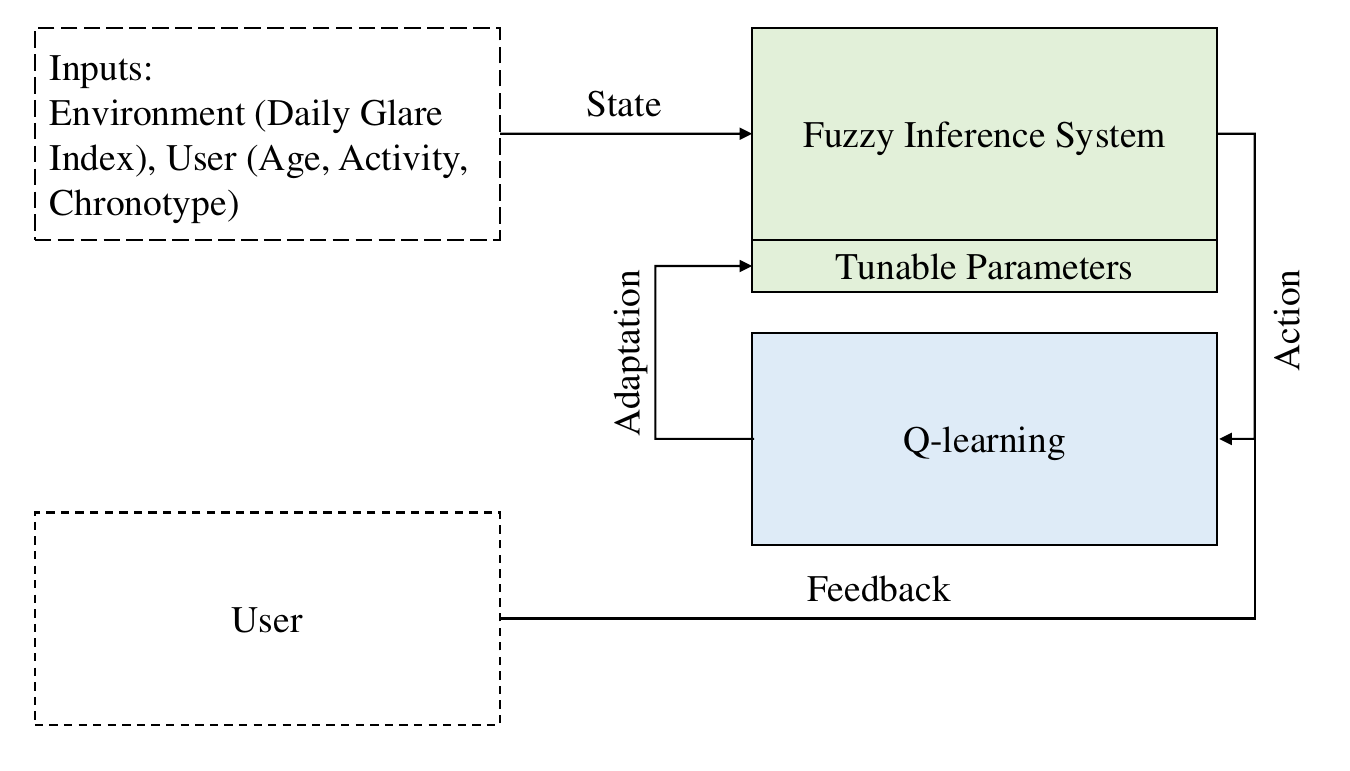}
    \caption{Architecture of the proposed QFIS algorithm for intelligent lighting management}
    \label{fig:blockdiagram}
\end{figure}

This section provides an overview of the proposed QFIS algorithm for intelligent lighting management.
Figure \ref{fig:blockdiagram} illustrates the algorithm architecture.
The inputs are categorized into three groups: (1) environment, (2) passenger, and (3) feedback for the RL agent.
For the environmental information, we use photopic sensors to measure the current ambient lighting quantified in terms of the daily glare index (DGI).
For the passenger information, we use age, chronotype, and the passenger's current activity. 
For any RL task, a mechanism for interaction with the environment is required.
For this purpose, we use a control knob by which the passenger can correct the lighting setting at any time.
The interactions with the control knob will constitute the reward feedback for the RL agent.
In this approach, if the algorithm changes lighting conditions and the passenger manually overrides it via the control knob, the adjustment applied by the passenger constitutes a negative reward that will be used for learning and adaptation.

Note that the selection of the above inputs was based on recommendations from our aircraft manufacturer partner.
In practice, the passenger age and chronotype can be recorded in a user-specific account within the aircraft infotainment system.
Subsequently, the passenger interactions with the control knob can be recorded in this account for continuous learning and adaptation over multiple flights.
Also, passenger activity can be obtained using a camera and various algorithms available in the vast literature of vision-based activity recognition \cite{zhang2017review}.

To generate an appropriate light setting, the environmental and passenger information enters the FIS module.
The role of FIS is to control lighting based on fuzzy MFs and rule-base that are developed based on expert knowledge.
Once FIS generates the light setting, the RL agent will monitor the passenger interactions with the control knob.
The passenger adjustments will be used to adapt FIS parameters according to the passenger preferences.
As will be described shortly, the adaptable parameters will modify the shape of MFs and the weights of rules in the fuzzy inference step.

The next two sections provide the details of the FIS and RL modules.

\subsection{Fuzzy inference system}\label{se:FIS}
Here, we explain the details of the proposed FIS in our algorithm.
In the design of the FIS, we use the Gaussian MFs and the Takagi-Sugeno (TS) inference system.
These choices ensure the smoothness and stability of the algorithm as it is detailed in  \cite{desouky2010learning}.
The inputs include age, activity, DGI, and chronotype. 
The output is the intensity of interior lights.
The structure of the fuzzy system is comprised of these components: fuzzification, rule-base, inference engine, and defuzzification, as detailed below.

\subsection{Fuzzification}
\begin{figure}[t]
    \centering
    \begin{subfigure}{0.45\textwidth}
        \centering
        \includegraphics[width=0.75\textwidth]{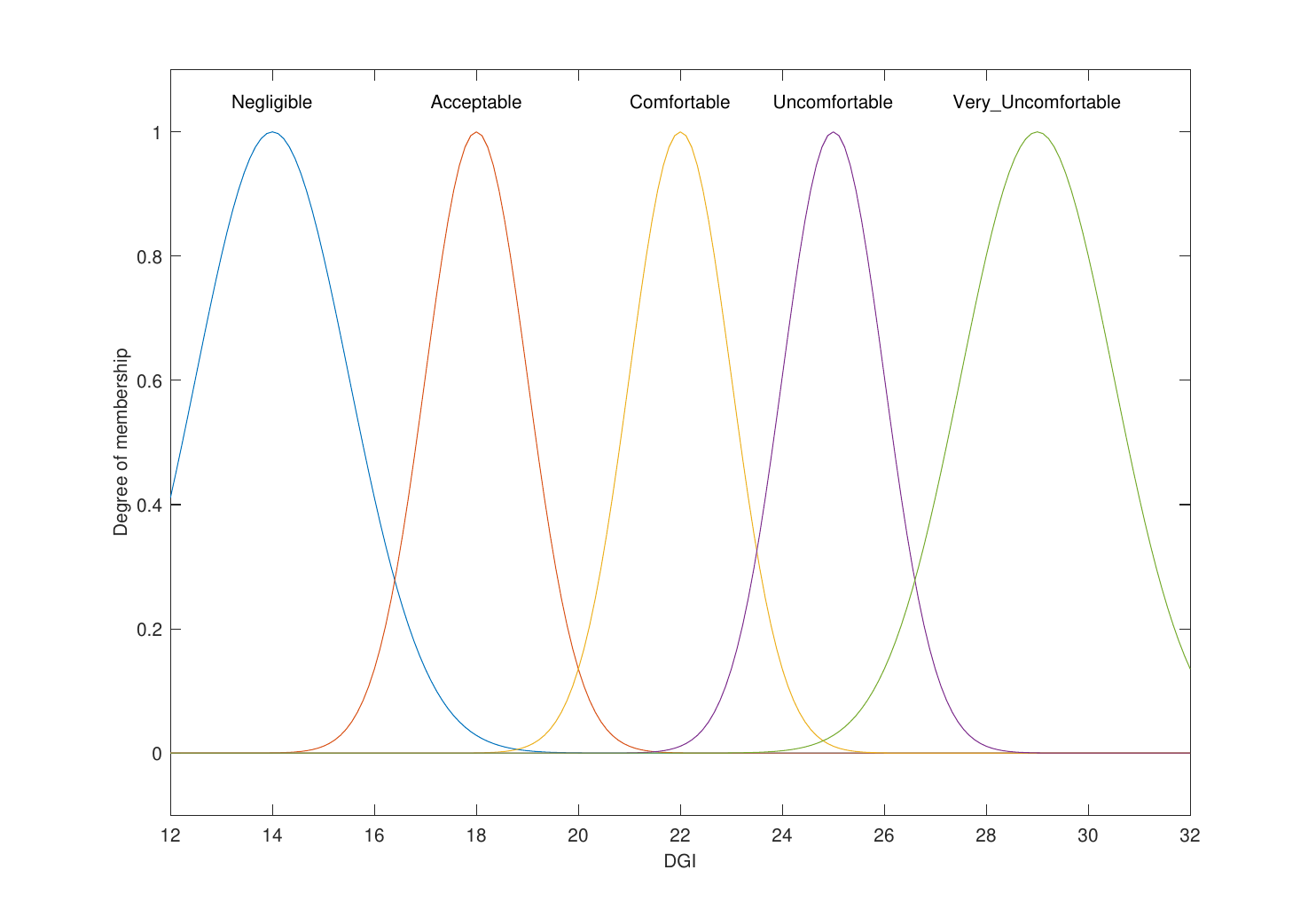}
        \caption{DGI}
        \label{fig:dgi}
    \end{subfigure}
    \hfill
    \begin{subfigure}{0.45\textwidth}
        \centering
        \includegraphics[width=0.75\textwidth]{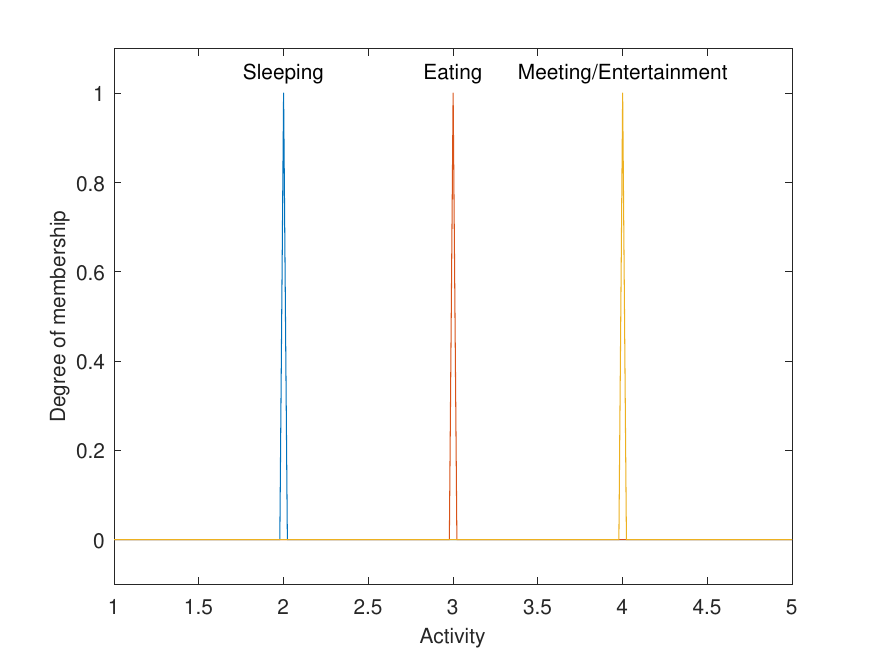}
        \caption{Activity}
        \label{fig:activity}
    \end{subfigure}
    \vspace{0.5cm}
    \begin{subfigure}{0.45\textwidth}
        \centering
        \includegraphics[width=0.75\textwidth]{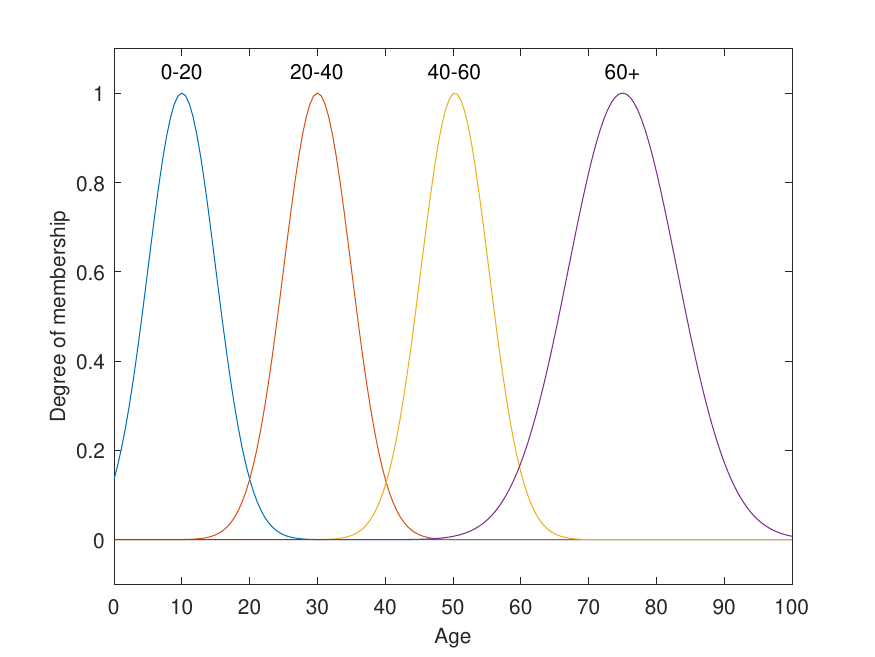}
        \caption{Age}
        \label{fig:age}
    \end{subfigure}
    \hfill
    \begin{subfigure}{0.45\textwidth}
        \centering
        \includegraphics[width=0.75\textwidth]{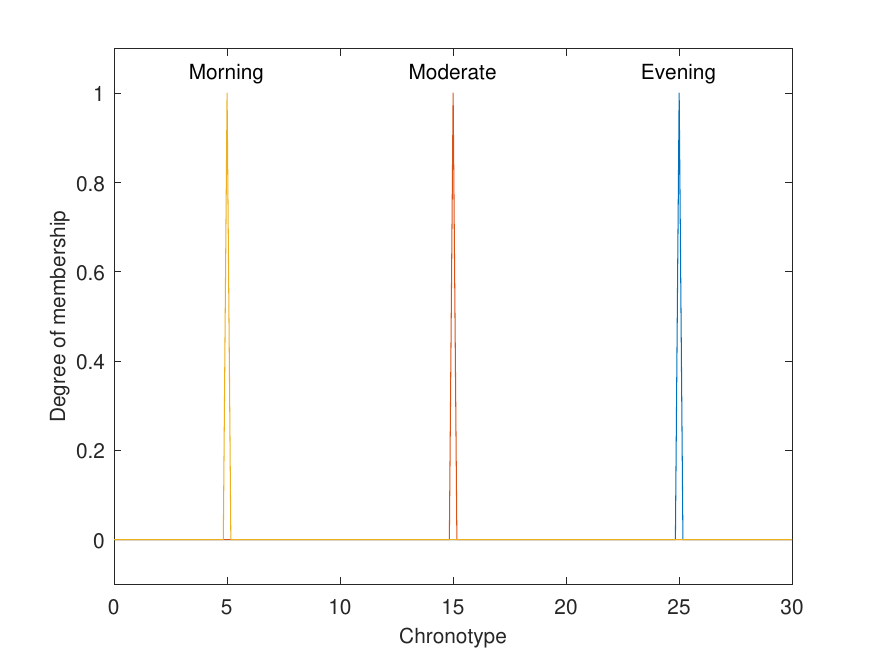}
        \caption{Chronotype}
        \label{fig:chronotype}
    \end{subfigure}
    \caption{MFs designed for each input of the FIS module}
    \label{fig:fuzzification}
\end{figure}

We use the Gaussian MF for the fuzzification of all inputs.
Let $m$ and $\sigma$ denote the mean and standard deviation of a Gaussian MF, then its output is expressed as
\begin{equation}\label{eq:Guassain}
    \mu\left(x\right)=\exp{\left(-\frac{1}{2}\left(\frac{x-m}{\sigma}\right)^2\right)}
\end{equation}
Figure \ref{fig:fuzzification} illustrates the MFs designed for each input.
While DGI and age are numerical inputs, passenger activity and chronotype are categorical.
To process all inputs in the same framework, we consider MFs with zero standard deviation for the categorical data and pass them through the fuzzification step.
The details of fuzzification for each input are given below.

\subsubsection{Daily glare index}
DGI is a common metric used to assess the potential discomfort caused by glare from daylight sources in indoor spaces. 
According to \cite{piccolo2009effect}, DGI values up to 22 are generally deemed acceptable, and higher values tend to become uncomfortable and even intolerable.
As such, we set the DGI value of 22 as a point of symmetry for the fuzzification of DGI, and use the deviations from this value to categorize the level of comfort/discomfort of ambient light (Fig. \ref{fig:dgi}).
Note that we overlap the MFs by 15 percent to ensure smooth transitions from one MF to another.
Table \ref{tab:dgi} shows the parameter values of DGI MFs.

\begin{table}[h]
\centering
\caption{Parameters of DGI MFs}
\small
\label{tab:dgi}
\begin{tabular}{lll}
\toprule
MF         & $m$ & $\sigma$ \\
\midrule
Negligible         & 14   & 1.5\\
Acceptable         & 18   & 1\\
Comfortable        & 22   & 1\\
Uncomfortable      & 25   & 1\\
Very Uncomfortable & 29   & 1.8\\
\bottomrule
\end{tabular}
\end{table}

\subsubsection{Activity}
We consider three types of activities: meeting/entertainment, eating, and sleeping. 
As the activity is categorical data, we use MFs with zero standard deviations and rank them based on their light requirements from low to high.
As shown in Fig. \ref{fig:activity}, we consider the same level of light intensity for meeting and entertainment and rank them higher than eating which itself ranks higher than sleeping.
As mentioned above, passenger activity is recognized by a computer vision algorithm, and the type of activities to be considered in our algorithm can be easily extended to an arbitrary number of activity categories.

\subsubsection{Age}
For the fuzzification of age, we rely on the findings of \cite{facchin2017glare} in which the glare thresholds for two major age groups, 20-40 and 40-60 years old, were studied. 
The luminosity thresholds for the 40-60 years old category are higher than those for the 20-40 years group. This information is helpful both in the design of the MFs and in setting up the rule base.
The MFs for age are illustrated in Fig. \ref{fig:age}, and the details of their parameters are presented in Tab. \ref{tab:age}.
Note that, aside from the 60+ category, we used identical standard deviations to ensure all age groups have an equal range of 20 years 
We also use a five percent overlap between MFs to handle transitions between the age brackets.

\subsubsection{Chronotype}
Chronotype refers to the inherent preference that individuals have for certain times of the day. 
It is a concept used to describe an individual's natural sleep-wake patterns and their corresponding preferences for being active or resting during particular parts of the day. 
Therefore, it can be used to tailor light settings to the passengers' preferences.

There are three primary chronotypes: morning, evening, and intermediate.
Morning chronotypes tend to feel more alert, awake, and productive during the early morning hours. 
Evening chronotypes feel more energetic, alert, and creative during the evening and nighttime hours. 
Intermediate chronotypes fall in between morning and evening chronotypes. 
The relationship between chronotypes and light exposure is well-studied in the literature \cite{martin2012relationship, figueiro2014effects, porcheret2018chronotype}.
It is shown that generally morning types prefer morning light exposure while evening types prefer evening light exposure. 
This allows us to rank the morning chronotype $>$ intermediate chronotype $>$ evening chronotype in terms of the tolerance of glare discomfort, leading to the MFs depicted in Fig. \ref{fig:chronotype}.

\begin{table}[t]
\centering
\caption{Parameters of age MFs}
\small
\label{tab:age}
\begin{tabular}{lll}
\toprule
MF         & $m$ & $\sigma$ \\\midrule
0-20 years old     & 10   & 5                  \\
20-40 years old    & 30   & 5                  \\
40-60 years old    & 50   & 5                  \\
60+ years old      & 75   & 8                  \\            
\bottomrule
\end{tabular}
\end{table}
\subsection{Rule-base}
With the above selection of MFs, there are five and four MFs allocated to DGI and age, respectively, while passenger activity and chronotype each have three MFs.
Therefore, there exist $5 \times 4 \times 3 \times 3 = 180$ possible combinations of fuzzy sets that must be taken into account in constructing the rule base.
The rules are created to cover all 180 combinations of inputs of the FIS with each rule corresponding to one of the fuzzy outputs. Due to the large number of rules, we present them in a supplementary document.

\subsection{Inference engine and defuzzification}
We use the zero-order TS inference engine for the design of the FIS.
As mentioned earlier, this choice ensures the stability of the QFIS algorithm as shown in \cite{desouky2010learning}.
With this choice, the FIS output will be a weighted average of the output values from the rules.

To formulate the FIS, let ${{\bf{x}}_t} = {\left[ {{x_{1,t}},{x_{2,t}},{x_{3,t}},{x_{4,t}}} \right]^T}$ be the vector of FIS inputs at time $t$. The $j$-th rule denoted by $R_j$ takes the following form
\begin{equation}
{R_j} = {\rm{IF}}\;\;\;{x_{1,t}}\;{\rm{is}}\;A_1^j\;\;\;{\rm{AND}}\;\;\; \cdots \;\;\; {\rm{AND}}\;\;\;{x_{4,t}}\;{\rm{is}}\;A_4^j\;\;\;{\rm{THEN}}\;\;\;{f_j}\left( {{{\bf{x}}_t}} \right) = {k_j},
\end{equation}
where $A_i^j$ is the fuzzy set for $x_i,t$, and the $k_j$ is the consequent parameter.
Assuming a weight of $w_j$ for $R_j$, the the output of the FIS at time $t$ will take the form
\begin{equation}\label{eq:fisOutput}
f\left( {{{\bf{x}}_t}} \right) = \sum\limits_{j = 1}^N {{{\bar w}_j}{k_j}} 
\end{equation}
where ${{\bar w}_j} = {w_j}{\left( {\sum\limits_{j = 1}^N {{w_j}} } \right)^{ - 1}}$ is the normalized weight, and $N=180$ is the number of rules.

\begin{figure}[h]
    \begin{subfigure}{0.5\textwidth} 
        \centering
        \includegraphics[width=0.6\linewidth]{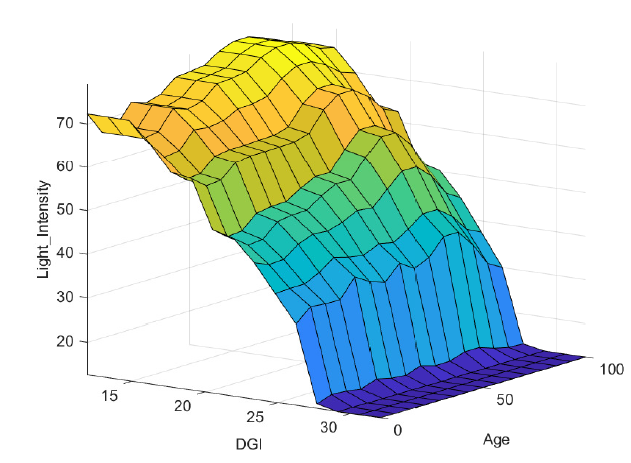}
    \end{subfigure}%
    \begin{subfigure}{0.5\textwidth} 
        \centering
        \includegraphics[width=0.6\linewidth]{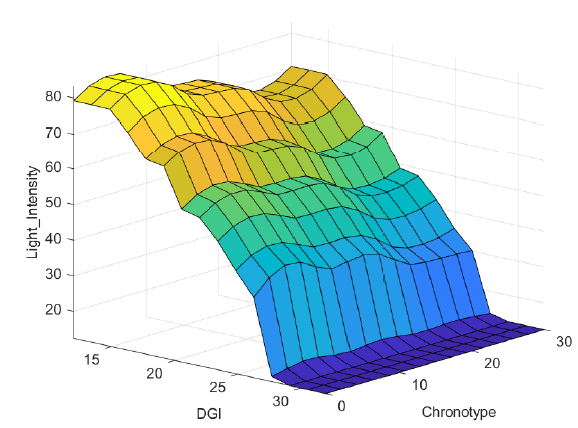}
    \end{subfigure}
    \caption{Fuzzy surface diagrams for the proposed FIS module}
    \label{fig:SurfaceDiagram}
\end{figure}

In the proposed algorithm, we make $k_j$s adaptable using Q-learning.
Another set of adaptable parameters are the mean values of MFs to be discussed later.
For the baseline FIS, we will use the $k_j$ values presented in Tab. \ref{tab:defuzzy}.
The weights associated with each rule represent the degree of activation or membership of the input variables in the corresponding fuzzy sets.
The weights will be kept constant during Q-learning adaptation.
To provide an insight into the characteristics of the proposed FIS, we present the fuzzy surface plots for DGI and chronotype, and DGI and age versus light intensity in Fig. \ref{fig:SurfaceDiagram}.

\begin{table}[h]
\centering
\caption{Output values chosen for the zero-order TS FIS}
\small
\label{tab:defuzzy}
\begin{tabular}{ll}
\toprule
FIS output category & $k$   \\ \midrule
Lights Off (D5)                                   & 0                              \\
Darken 4 (D4)                                    & 12.5                           \\
Darken 3 (D3)                                    & 25                             \\
Darken 2 (D2)                                    & 37.5                           \\
Darken 1 (D1)                                   & 50                             \\
Light up 1 (LU1)                                   & 62.5                           \\
Light up 2 (LU2)                                 & 75                             \\
Light up 3 (LU3)                                  & 87.5                           \\
Light up 4 (LU4)                                   & 100                            \\ \bottomrule 
\end{tabular}
\end{table}

\section{Reinforcement learning}\label{se:algorithm}
In this section, we augment the aforementioned FIS with RL, to create the QFIS algorithm with learning and adaptation capabilities.
As discussed earlier, given the relatively small, discrete, and model-free characteristics of the problem domain, we will base our approach on Q-learning.
Similar to conventional Q-learning algorithms, we represent the state space using Q tables.

Let ${{\bf{Q}}_t}$ be the Q-table at time $t$.
Given the state ${{\bf{x}}_t} = {\left[ {{x_{1t}},{x_{2t}},{x_{3t}},{x_{4t}}} \right]^T} \in {{\bf{Q}}_t}$, the FIS module takes the action $f\left( {{{\bf{x}}_t}} \right)$, leading to a reward ${r_t}$ that is defined using the following proposed reward function
 \begin{equation}\label{eq:rewardFunction}
r_t=-\frac{2}{\pi} \arctan \left(\epsilon(f\left( {{{\bf{x}}_t}} \right)-a)\right),
\end{equation}
where $\epsilon$ is a bias value that suggests how harsh or forgiving the reward function is, and $a$ represents the desired light intensity change as determined by user preferences.

Using $r_t$, we will update the Q-table as follows  
\begin{equation}\label{eq:Qlearning}
{{\bf{Q}}_{t + 1}}\left( {{{\bf{x}}_t},\;f\left( {{{\bf{x}}_t}} \right)} \right) = \;{{\bf{Q}}_t}\left( {{{\bf{x}}_t},\;f\left( {{{\bf{x}}_t}} \right)} \right) + \;\eta {\Delta _t}
\end{equation}
where $\eta > 0$ is the learning rate, and $\Delta_t$ is the temporal difference error defined as follows
\begin{equation}\label{eq:temporaldifference}
{\Delta _t} = {r_t}\left( {{{\bf{x}}_t},\;f\left( {{{\bf{x}}_t}} \right)} \right) + \;\gamma {V_t}({{\bf{x}}_t}) - \;{{\bf{Q}}_t}\left( {{{\bf{x}}_t},\;f\left( {{{\bf{x}}_t}} \right)} \right),
\end{equation}
where $\gamma > 0$ is known as the discount factor and ${V_t}({{\bf{x}}_t}) = \mathop {\max }\limits_{f\left( {{{\bf{x}}_t}} \right)} \;{{\bf{Q}}_t}\left( {{{\bf{x}}_t},\;f\left( {{{\bf{x}}_t}} \right)} \right)$.

To adapt the FIS module to the user preferences, we construct adaptation rules for the MF mean values $m$ and the output values $k$.
Let $\bf{m}$ and $\bf{k}$ be two vectors consisting of all MFs' mean and FIS output values to be tuned.
Then, the vector of all adaptable parameters is $\boldsymbol{\phi} = {[{{\bf{m}}^T}\;{{\bf{k}}^T}]^T}$.
To derive adaptation laws for these parameters, let us define the following objective function
\begin{equation}
    E = \frac{1}{2}\Delta_t^2.
\end{equation}
Using the gradient decent approach, the parameters can be updated as follows
\begin{equation}
    \phi \left( {t + 1} \right) = \phi \left( t \right) - \eta \frac{{\partial E}}{{\partial \boldsymbol{\phi} }}
\end{equation}
We note that
\begin{equation}\label{eq:partial}
\frac{{\partial E}}{{\partial {\boldsymbol{\phi}}}} = {\Delta _t}\frac{{\partial {\Delta _t}}}{{\partial {\boldsymbol{\phi}}}}.
\end{equation}
Substituting \eqref{eq:temporaldifference} into \eqref{eq:partial} yields
\begin{equation}
    \frac{{\partial E}}{{\partial {\boldsymbol{\phi}}}} = -{\Delta _t}\frac{{\partial {Q_t}\left( {{{\bf{x}}_t},f({{\bf{x}}_t})} \right)}}{{\partial {\boldsymbol{\phi}}}}
\end{equation}
Using \eqref{eq:fisOutput}, we have
\begin{equation}\label{eq:updateRulem}
    \frac{{\partial {{\bf{Q}}_t}\left( {{{\bf{x}}_t},f({{\bf{x}}_t})} \right)}}{{\partial {k_j}}} = {\bar w_j},
\end{equation}
where $j$ corresponds to the $j-$th rule. Moreover, we have
\begin{equation}
\frac{{\partial {{\bf{Q}}_t}\left( {{{\bf{x}}_t},f({{\bf{x}}_t})} \right)}}{{\partial m_i^j}} = \frac{{\partial {{\bf{Q}}_t}\left( {{{\bf{x}}_t},f({{\bf{x}}_t})} \right)}}{{\partial {w_j}}}\frac{{\partial {w_j}}}{{\partial m_i^j}}\; = \frac{{{k_j} - {{\bf{Q}}_t}\left( {{{\bf{x}}_t},f({{\bf{x}}_t})} \right)}}{{\sum\nolimits_j {{w_j}} }}{w_j}\frac{{{x_i} - m_i^j}}{{{{\left( {\sigma _i^j} \right)}^2}}},
\end{equation}
where the superscript refers to the $j$-th rule.
Note that in driving \eqref{eq:updateRulem}, we use \eqref{eq:Guassain}.
With the above developments, the parameters update rule can be summarized as
\begin{equation}\label{eq:4.14}
\left[ {\begin{array}{*{20}{c}}
{m_{i,t + 1}^j}\\
{{k_{j,t + 1}}}
\end{array}} \right] = \left[ {\begin{array}{*{20}{c}}
{m_{i,t}^j}\\
{{k_{j,t}}}
\end{array}} \right] + \eta {\Delta _t}\left[ {\begin{array}{*{20}{c}}
{\frac{{\left( {{k_j} - {{\bf{Q}}_t}\left( {{{\bf{x}}_t},f({{\bf{x}}_t})} \right)} \right)}}{{\sum\nolimits_j {{w_j}} }}{w_j}\frac{{{x_i} - m_i^j}}{{{{\left( {\sigma _i^j} \right)}^2}}}}\\
{{{\bar w}_j}}
\end{array}} \right].
\end{equation}

\subsection{Q-table formation}
For our intelligent light management algorithm, three scenarios are possible, including when the FIS output is: (1) correct i.e. it is the same as the user's preference, (2) too bright, and a negative reward is given to lower the intensity levels, and (c) too dark, and a negative reward is given to raise the intensity levels.
Here, a single Q-table will be ineffective as it can not differentiate between two negative reward scenarios.
Therefore, we use two Q-tables as follows to account for both scenarios.
This way, we will be able to appropriately reward the algorithm for suggesting lighting intensities that are either too bright or too dark.
The combinations of all four inputs of the algorithm and their corresponding MFs create 180 different states in the Q-tables.

\section{Implementation and results}\label{se:Simulations}
We conducted an extensive user study in an aircraft cabin mockup to assess the performance of the proposed algorithm. 
This section details our experimental setup followed by the test procedures and their results.

\subsection{Experiments setup}
\begin{figure}[h]
    \centering
    \includegraphics[width=0.72\textwidth]{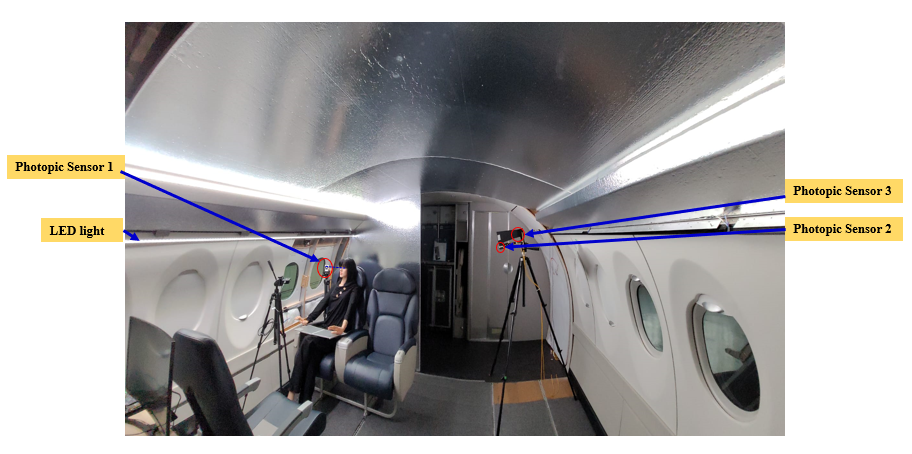}
    \caption{Aircraft cabin mockup for the experiments}\label{fig:cabin}
    \centering
    \includegraphics[width=0.52\textwidth]{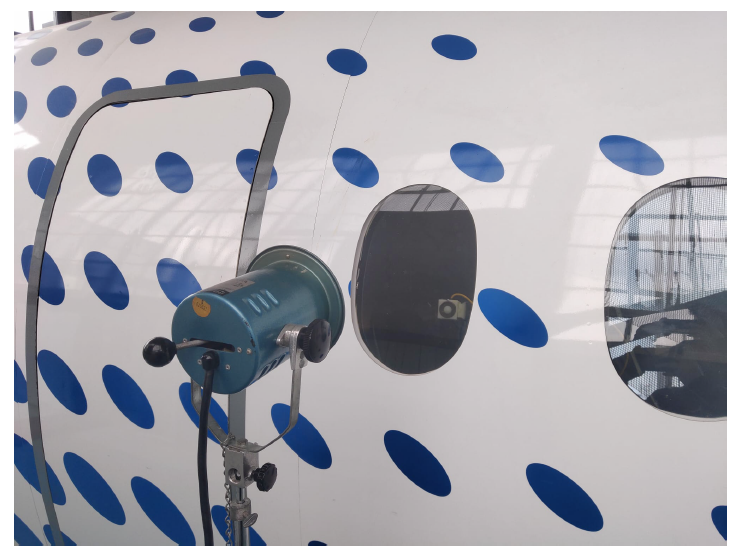}
    \caption{The setup for simulating exterior lighting}\label{fig:outsidethecabin}
\end{figure}

For this purpose, we established an experimental setup using a real aircraft fuselage within a research laboratory at Toronto Metropolitan University, as depicted in Fig. \ref{fig:cabin}.
To emulate the effect of external sunlight inside the cabin, we positioned a 1000W Colortran LQF6 floodlight, featuring an output of roughly 57000 lm and a color temperature of 5000K, outside the aircraft fuselage. This setup can be seen in Fig. \ref{fig:outsidethecabin}. 

\begin{figure}[h]
    \centering
    \includegraphics[width=0.6\textwidth]{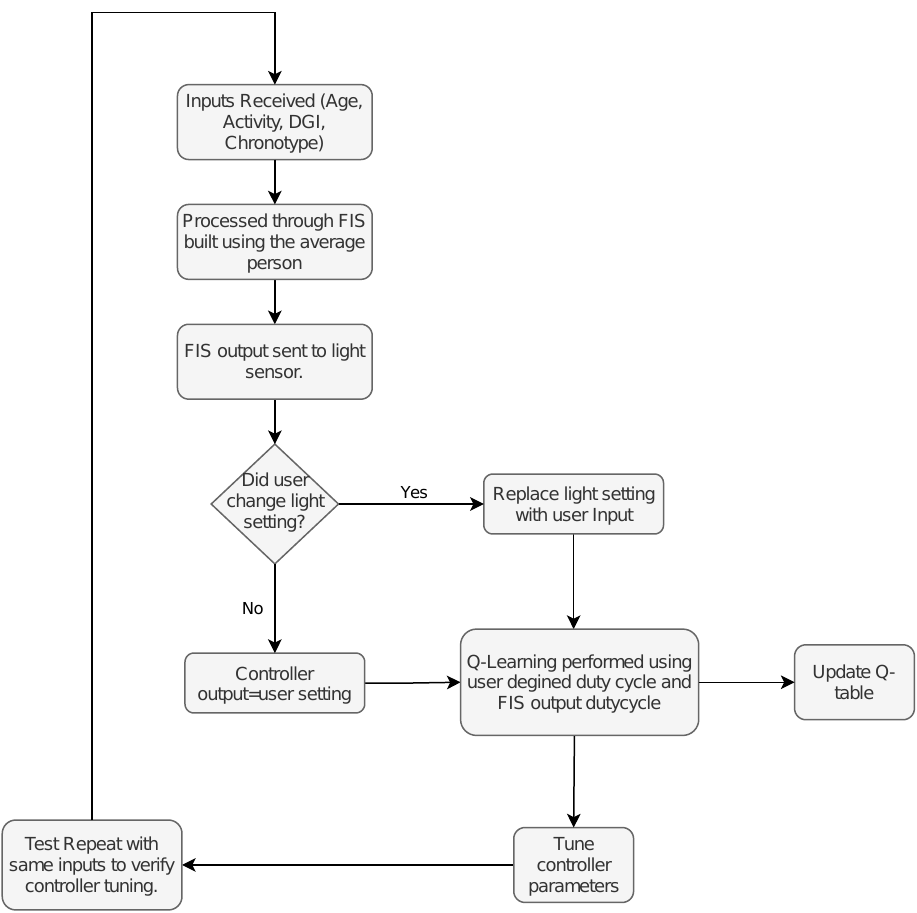}
    \caption{Flowchart of experimental procedure}\label{fig:flowchart}
\end{figure}

Inside the cabin, illumination was provided by an intelligent LED strip, producing 1600 lm, which was controlled via a Raspberry Pi. We employed a photopic sensor (PMA 1130-S-420-150K) procured from Solar Light Company, Glenside, PA, to measure the Direct Glare Index (DGI).
The backbone of our system's computation was the NVIDIA Jetson TX2, operating on a Linux system. To make the testing process more efficient and enhance user engagement, we developed a mobile application. Using this platform, participants were able to transmit specific inputs, including activity, age, and chronotype, directly to our algorithm.
The FIS utilized these inputs to determine the light intensity output, which was subsequently conveyed to the LED strip in the form of a duty cycle. Participants were then exposed to the resulting lighting conditions. They were given the flexibility to either adjust the light intensity through the application or continue with the preset configurations.
Such feedback became instrumental for the Q-learning algorithm, enabling it to optimize the parameters of the FIS. Figure \ref{fig:flowchart} presents a comprehensive outline of our experimental procedures.

We engaged 10 participants from the 20-40 age bracket and 8 from the 40-60 age bracket. 
These age groups were recommended by our industry partner, as most passengers, especially for business jets, belong to these two age groups.

\subsection{FIS results}
Before we test the proposed QFIS algorithm, it is important to evaluate the FIS module of the algorithm to ensure it generates appropriate light settings for an average individual in each age group.
For this purpose, each participant was placed in a set scenario and asked to correct their preference for light intensity, using the aforementioned mobile application.
The experiments encompassed three distinct activities, including entertainment, eating, and sleeping. 
Based on the recommendations given in \cite{ochoa2012considerations}, the target light settings for these activities include 450 lux for entertainment, 150-200 lux for eating, and less than 50 lux for sleeping.

\begin{figure}[h]
    \begin{subfigure}{0.495\textwidth}
    \centering
    \includegraphics[width=\linewidth]{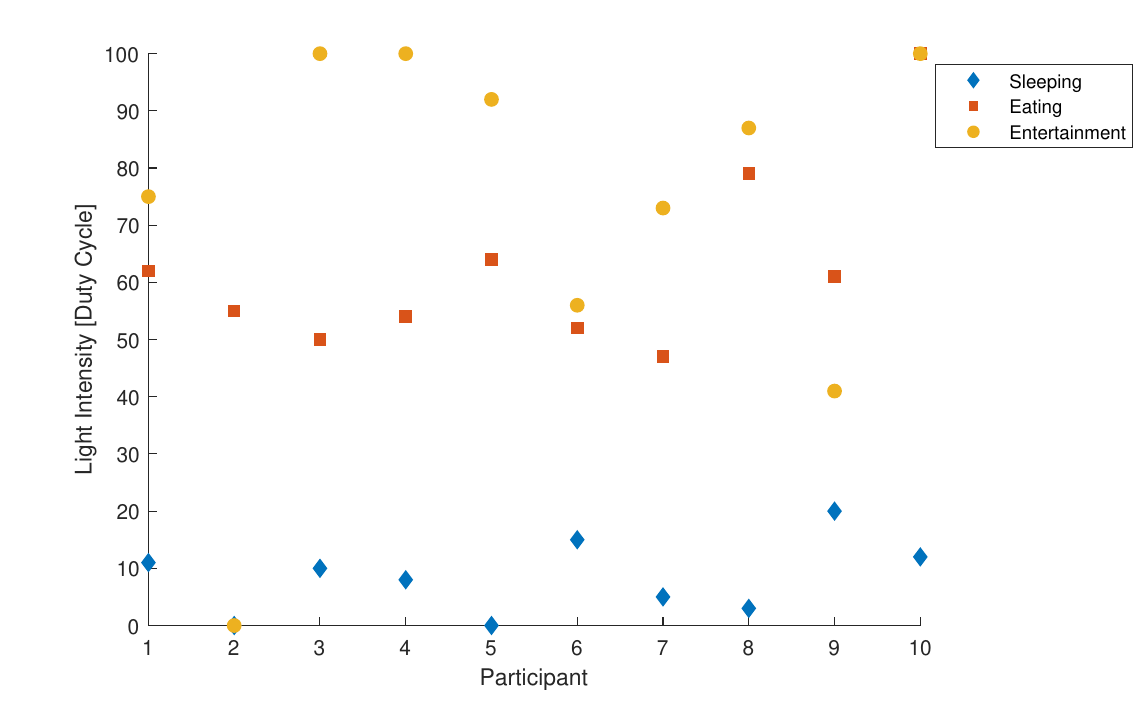}
    \end{subfigure}
    \begin{subfigure}{0.495\textwidth}
    \centering
    \includegraphics[width=\linewidth]{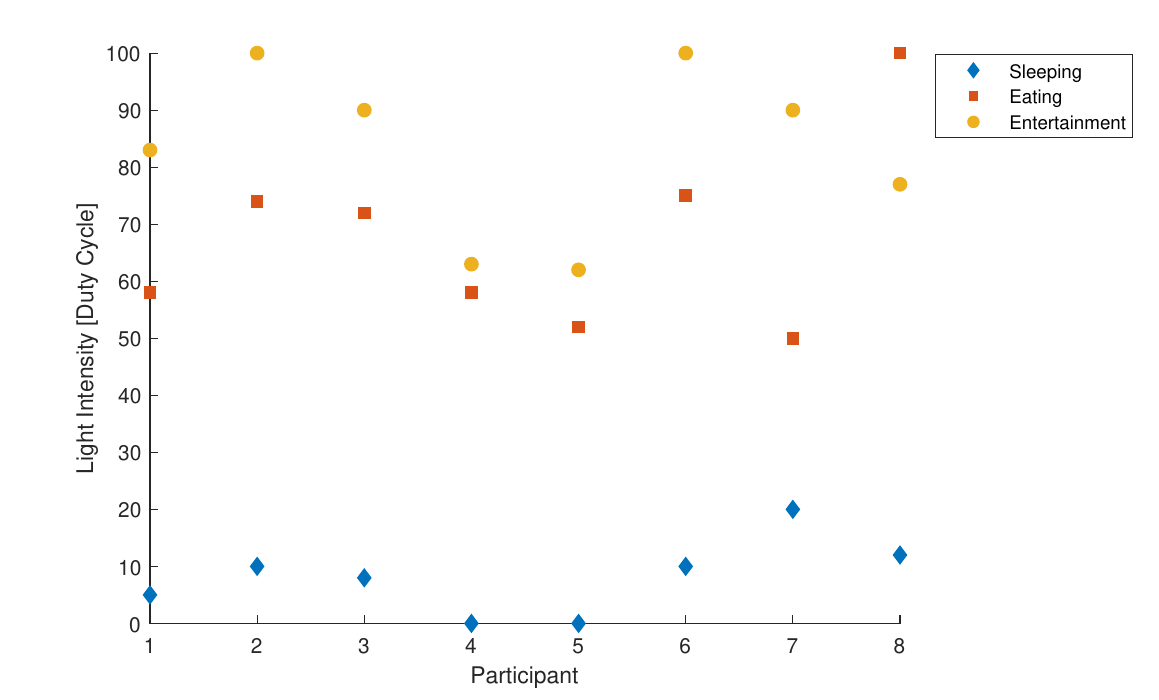}
    \end{subfigure}
    \caption{The baseline FIS light intensity output for the 20-40 years age group (left), and 40-60 years age group (right)}
    \label{fig:fisExperiments}
\end{figure}

Figure \ref{fig:fisExperiments} illustrates the light intensity prescribed by the FIS module for each participant across different activities for the two age group categories.
Following the completion of the trials, the results for each activity were averaged to obtain an understanding of the preferred average lighting settings in different activities and age ranges (Tab. \ref{tab:fisExperiments}).
This data highlights the difference between the light intensity comfort levels in different circumstances.
Additionally, this data plays an important role in understanding passengers whose preferences do not fall under the average patterns in each category.

\begin{table}[h]
\centering
\caption{The average of light intensity values generated by the baseline FIS module}
\small
\label{tab:fisExperiments}
\begin{tabular}{lll}
\toprule
Activity & 20 – 40 years age group & 40 - 60 years age group \\ \cmidrule{2-3}
Entertainment & 72.4 & 83.125 \\
Eating & 62.4 & 67.375 \\
Sleeping & 8.4 & 8.125 \\ \bottomrule
\end{tabular}
\end{table}

\subsection{QFIS Results}
This section explores the learning capabilities of the proposed algorithm. Detailed below are three different experimental sets, each encompassing participants from varying age groups with notably distinct lighting preferences and engaged in different activities:
\begin{enumerate}
\item \textit{Experiments set 1}, featuring a participant aged between 20-40 years who prefers significantly dimmer light settings relative to the typical preferences of this age group, while engaged in entertainment.
\item \textit{Experiments set 2}, including a participant aged between 40-60 years who leans towards much brighter light settings compared to the standard settings of this age range, while having a meal.
\item \textit{Experiments set 3}, focusing on a participant from the 20-40 years age bracket, while attempting to sleep.
\end{enumerate}


The success criteria for the algorithm constituted two factors: 
\begin{enumerate}
    \item The correctness of the tuning: Does the algorithm correctly tune the output towards the desired passenger light intensity?
    \item The effectiveness of the tuning: comparing different learning rates $\eta$ and the number of trials required to tune different scenarios.
\end{enumerate}

\subsubsection{Experiments set 1}
Table \ref{tab:paramsfirstscenario} presents the input state, the initial algorithm output, and the participant's preference in this set of experiments.

For the first test case, the objective was to quickly tune the $k$ parameters to align with user preferences.
The user used the aforementioned mobile application to interact with the algorithm and provide feedback.
The algorithm calculated a reward from the feedback at every iteration and adapted the $k$ parameter, leaning toward the user preference.
To explore the effects of varying learning rates, we evaluated learning rates from 0.1 to 0.5.

\begin{table}[t]
\centering
\caption{Parameters for the experiments set 1}
\small
\label{tab:paramsfirstscenario}
\begin{tabular}{llllll}
\hline
Variable             & Category      & Value    &  &  &  \\ \hline
Age                  & 20-40         & 22       &  &  &  \\
DGI                  & Comfortable   & 22       &  &  &  \\
Chronotype           & Evening     & 25       &  &  &  \\
Activity             & Entertainment & 5        &  &  &  \\
Baseline FIS output    & LU2           & 75       &  &  &  \\
Passenger Preference & $\sim$LU1     & $\sim$62 &  &  &  \\ \bottomrule
\end{tabular}
\end{table}

\begin{figure}[t]
    \centering
    \includegraphics[width=0.8\textwidth]{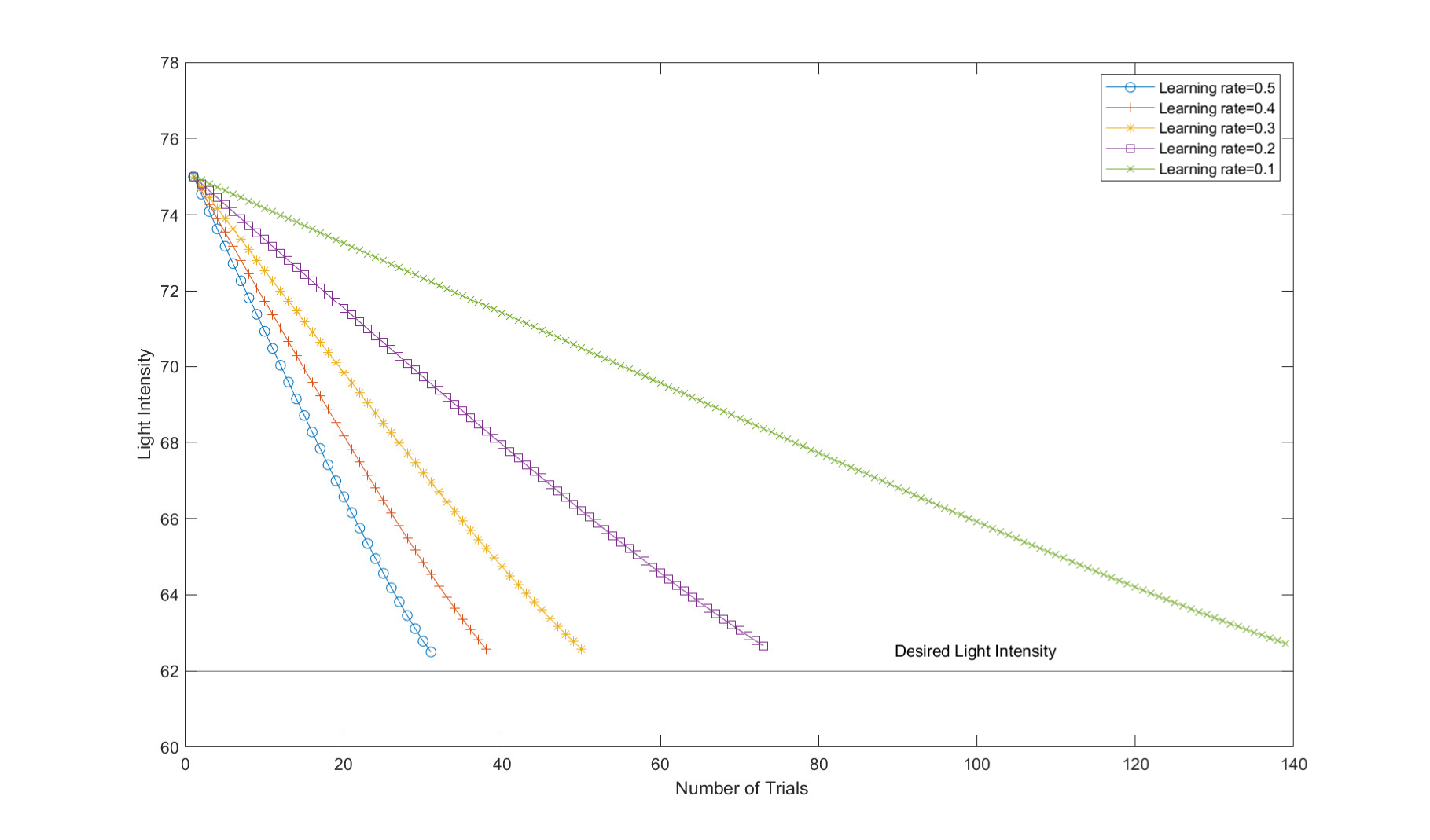}
    \caption{The change in the algorithm output with varying learning rates in experiments set 1}
    \label{fig:firstRLTest}
\end{figure}

Figure \ref{fig:firstRLTest} illustrates experimental results. 
While the algorithm starts with the initial output of 75, it learns from the user inputs and adapts its parameters to reach the user preferences over time.
It is evident that the value of $\eta$ dramatically influences the number of trials necessary.
For instance, when $\eta=0.5$, the algorithm tuned with approximately 22 percent less number of trials compared to the case of $\eta=0.1$.
This highlights that by choosing an appropriate value for $\eta$, one can effectively control the learning behavior of the algorithm.

Exploring the variations in the $m$ parameters during these trials revealed they behaved in three distinct manners:
\begin{enumerate}
\item When the input directly corresponds to the mean of MF, the mean remains unchanged because it represents the desired value for the input.
\item If the input is positioned closely between two MFs, the specific state oscillates between these two MFs, thus fine-tuning both $k$ parameters concurrently.
\item When the input predominantly aligns with one MF, the mean of that function undergoes sole adjustment to expedite the tuning, ensuring the output reflects the passenger's preference.
\end{enumerate}

To further explore the role of $m$ parameters, we conducted additional experiments, observing the $m$ values during prolonged trials.
We set $\eta=0.1$ for tuning all $m$ and $k$ parameters.
Table \ref{tab:secondScenario} shows the final values of $m$ for the age MFs and their corresponding $k$ values.

\begin{table}[t]
\centering
\caption{The shift in $m$ values with different learning rates in experiments set 1}
\small
\label{tab:secondScenario}
\begin{tabular}{lll}
\toprule
Number of trials & $\eta$ & $m$ \\ \midrule
48  & 0.5 & 30 to 27.732 \\
59  & 0.4 & 30 to 27.409  \\
71  & 0.3 & 30 to 27.995\\
87  & 0.2 & 30 to 27.498   \\
136 & 0.1 & 30 to 27.066   \\ \bottomrule
\end{tabular}
\end{table}

To elaborate on the results, recall that while this particular participant belongs to the age group 20 - 40 years old, their lighting preference aligns more with a younger age bracket.
Therefore, the algorithm struggles to identify an optimal MF for the user.
As $\eta$ of all $m$ and $k$ parameters are identical, the algorithm tunes all the parameters simultaneously to generate the optimal output.
However, this leads to large shifts in MFs, ultimately centering them around either 10 or 34. 
While this adaptation finally results in an output aligned with the user preference, a significant drawback is the extended duration required to reach the desired performance.
This can be alleviated by setting a low learning rate (e.g., 0.002) for the $m$ parameters while maintaining a relatively larger learning rate for the $k$ parameters (e.g., 0.1).
With such choices of learning rates, $k$ parameters can quickly converge to a neighborhood of desired behavior, while slight changes in the $m$ values will fine-tune the algorithm behavior.

It is noteworthy that the DGI MFs were not affected in this test because the DGI in the test environment was 22, falling directly on the mean of the \textit{Comfortable} DGI MF. 

\subsubsection{Experiments set 2}
Table \ref{tab:paramssecondparticipant} presents the parameters for this set of experiments.
Note that this set of experiments involves a different level of DGI, and new participants with a different age category, chronotype, and activity compared to experiments set 1. 
Furthermore, this participant's lighting preferences align more with an older age bracket; therefore, the new experiments evaluate the algorithm in scenarios opposite to the ones mentioned in the previous section.

\begin{table}[t]
\centering
\caption{Parameters for the experiments set 2}
\small
\label{tab:paramssecondparticipant}
\begin{tabular}{lll}
\hline
Variable             & Category      & Value  \\ \hline
Age                  & 40-60         & 50   \\
DGI                  & Negligible   & 14        \\
Chronotype           & Morning     & 5       \\
Activity             & Eating & 3       \\
Baseline FIS output    & LU3           & 87.5      \\
Passenger Preference & $\sim$LU4     & $\sim$100  \\ \bottomrule
\end{tabular}
\end{table}

\begin{figure}[t]
    \centering
    \includegraphics[width=0.8\textwidth]{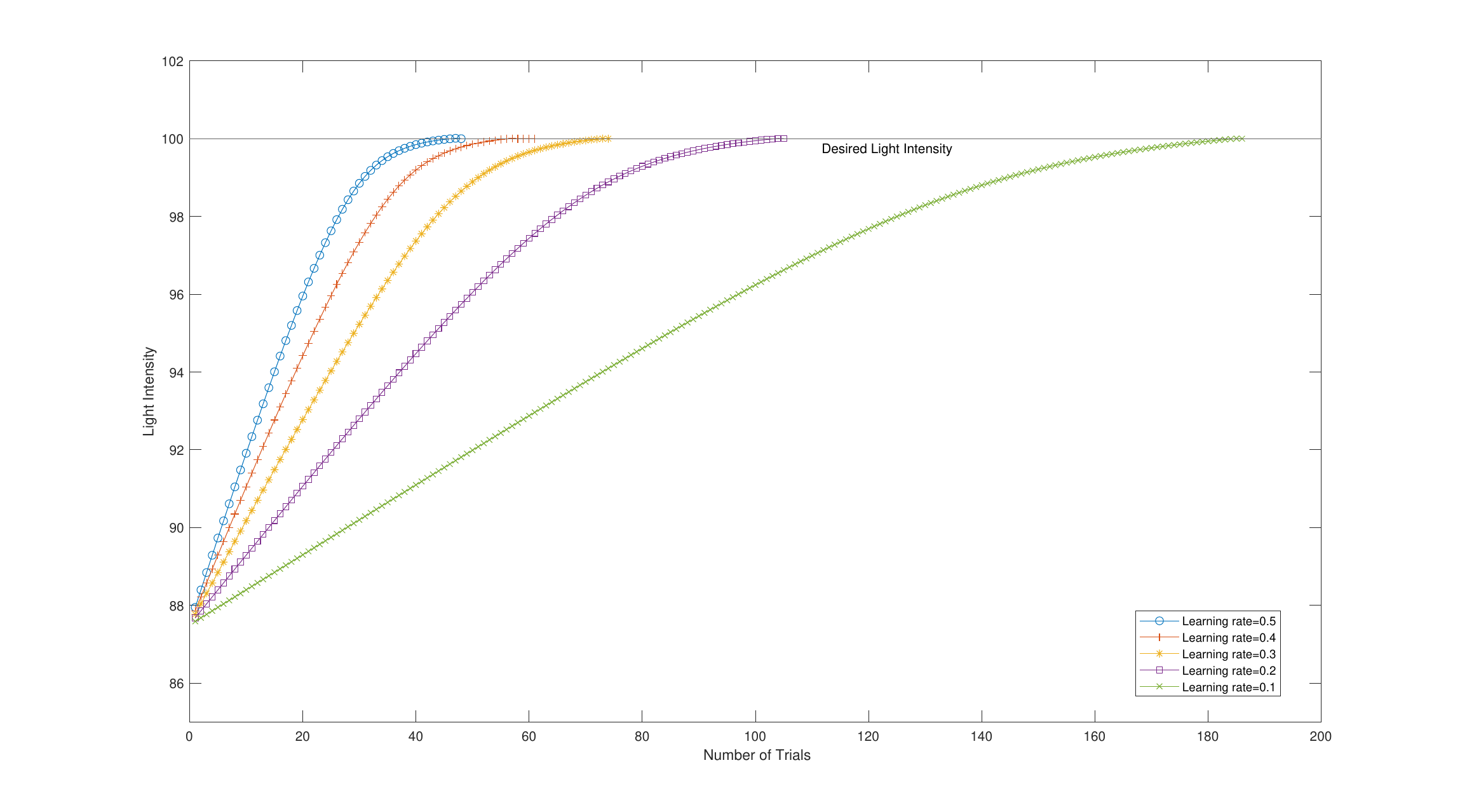}
    \caption{The change in the algorithm output with varying learning rates in experiments set 2}
    \label{fig:secondparticipant}
\end{figure}

We observed the learning behavior of the algorithm for different $\eta$ values ranging from 0.1 to 0.5.
Figure \ref{fig:secondparticipant} illustrates the results, showing that the algorithm can effectively learn the passenger preferences and adapt its parameters correctly for cases where the user prefers more lighting. Furthermore, when comparing the varying learning rates, using $\eta=0.5$ allowed for 25 percent faster tuning compared to the case of $\eta=0.1$.
These results, coupled with outcomes from experiments set 1, verify the algorithm's effectiveness in adapting to the preferences of users with varied characteristics.

Further, upon closely examining the algorithm's performance near the target behavior, it is evident that there is minimal or no overshoot at all. 
Table \ref{tab:overshoot1} presents the final three trials from experiments set 2 with $\eta=0.2$.
The algorithm exceeds the desired value of 100 by a marginal amount, but this leads to another negative reward that is accounted for by our dual Q-table configuration. 
Consequently, once the algorithm exceeds the desired value, it promptly readjusts to the target in the subsequent trial.
It is important to highlight that while higher learning rates might lead to more noticeable overshoots; these can be effectively handled with adaptive learning rates.

\begin{table}[t]
\centering
\caption{Algorithm overshoot and its correction}\label{tab:overshoot1}
\small
\begin{tabular}{llll}
\toprule
Trial number & $\eta$ & $r_t$ & $f\left({\bf{x}}_t\right)$ \\ \midrule
103               & 0.2           & -0.0254 &  99.99203       \\
104               & 0.2           & -0.0159  &  100.005    \\
105               & 0.2           & 0    &   100  \\ \bottomrule
\end{tabular}
\end{table}

\subsubsection{Experiments set 3}
This set of experiments focuses on sleeping, as it is a unique activity in which the interior lights need to shut off or function as a nightlight.
Through the FIS tests, it was found that the majority of individuals prefer a light intensity of around 10. However, to challenge the algorithm, we asked a participant who prefers brighter lighting for sleeping to conduct this set of experiments.
Table \ref{tab:thirdparticipant} shows the parameters for these trials.

\begin{table}[t]
\centering
\caption{Parameters for the experiments set 3}\label{tab:thirdparticipant}
\small
\begin{tabular}{llllll}
\toprule
Variable             & Category    & Value    &  &  &  \\ \midrule
Age                  & 20-40       & 27       &  &  &  \\
DGI                  & Comfortable & 22       &  &  &  \\
Chronotype           & Night   & 25       &  &  &  \\
Activity             & Sleeping    & 2        &  &  &  \\
Baseline FIS output    & D4          & 12.5     &  &  &  \\
Passenger Preference & D2          & 35 &  &  & \\ \bottomrule
\end{tabular}
\end{table}

\begin{figure}
    \centering
    \includegraphics[width=0.8\textwidth]{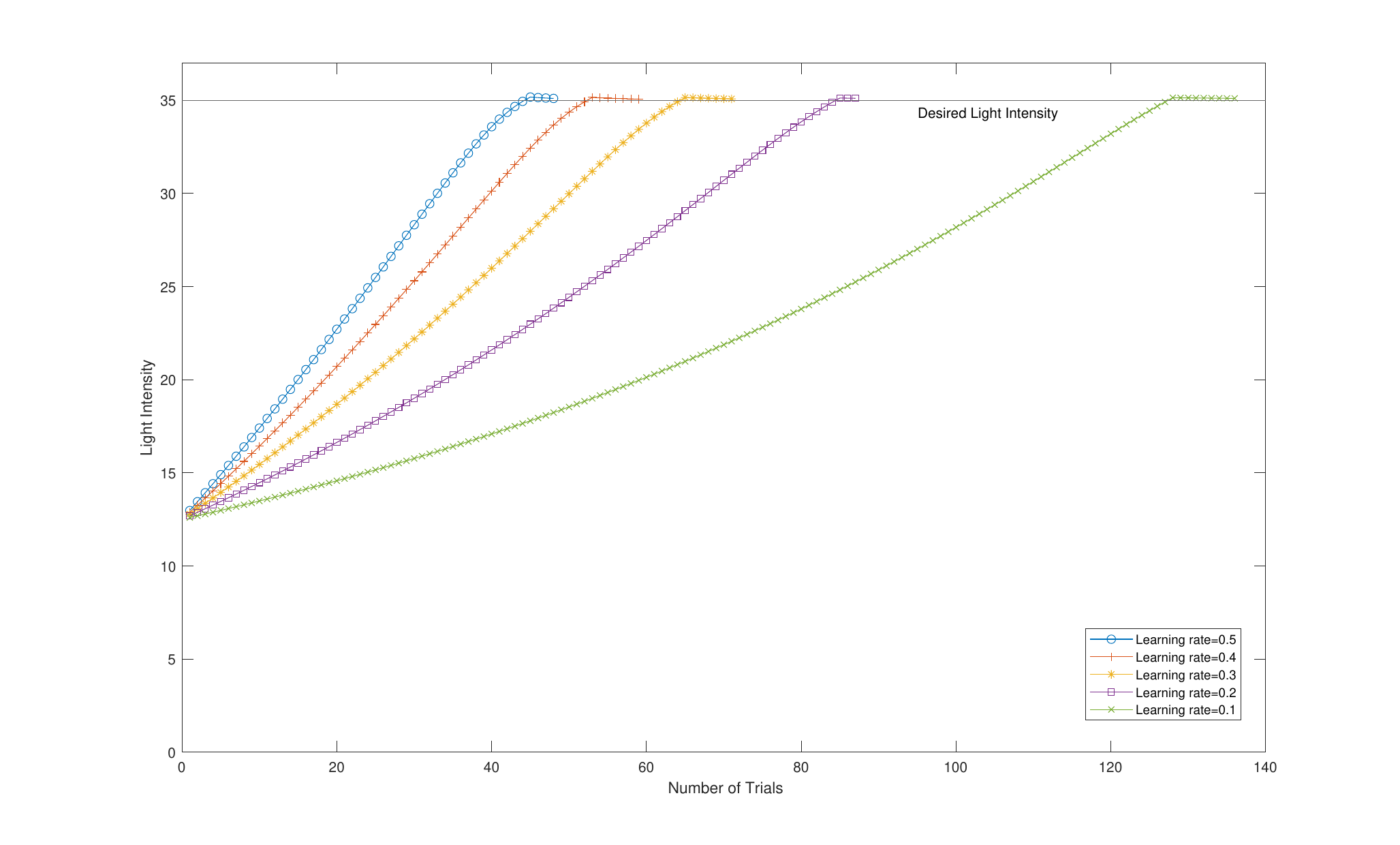}
    \caption{The change in the algorithm output with varying learning rates in experiments set 3}
    \label{fig:thirdparticipant}
\end{figure}

Figure \ref{fig:thirdparticipant} displays the test results highlighting the effects of various learning rates.
This test displays a higher range of tuning as the participant required a higher light intensity when compared to the baseline FIS's initial output.
Observing the behavior $m$ parameters, it was revealed that the user age predominantly aligned with one MF, and the mean of that MF was tuned solely to accelerate tuning to match the user preference.
This can be seen in Fig. \ref{fig:thirdparticipant} by the slight increase in the slope of the curves over 
Table \ref{tab:thirdparticiapantsecondtable} tabulates the shift of the mean from the first test to the last for different $\eta$ values.

\begin{table}[t]
\centering
\caption{The shift in $m$ values ith different learning rates in experiments set 3}\label{tab:thirdparticiapantsecondtable}
\small
\begin{tabular}{lll}
\toprule
Number of trials & $\eta$ & $m$ \\ \midrule
48               & 0.5           & 30 to 27.732         \\
59               & 0.4           & 30 to 27.409         \\
71               & 0.3           & 30 to 26.995         \\
87               & 0.2           & 30 to 26.498         \\
136              & 0.1           & 30 to 26.066         \\ \bottomrule
\end{tabular}
\end{table}

\subsection{Summary of results}
Through the various experiments conducted, the algorithm's effectiveness in adapting to users with a wide spectrum of preferences and characteristics was verified. Furthermore, the effect of various learning rates was explored, showing that the learning behavior of the algorithm can be controlled easily using the learning rate.

The detailed examination of the algorithm behavior revealed that the learning rate for the $m$ parameters should be smaller than the ones for the $k$ parameters, crucial for a balanced adaptation of algorithm parameters and minimizing the number of trials needed to match the user preferences.

The number of trials also depends on the difference between the baseline FIS output, and the passenger preference.
This underscores the importance of using the domain knowledge to build a fine-tuned baseline FIS.


\section{Conclusion}
This paper developed a QFIS algorithm for intelligent lighting management with a focus on aircraft interiors. Through a comprehensive user study, we have demonstrated the algorithm's efficacy in accommodating a diverse spectrum of user preferences and characteristics. Furthermore, we have conducted an in-depth analysis of the algorithm's behavior and studied the impact of its design parameters.

While our focus was on aircraft interiors, the developed algorithm can be used in any environment that can benefit from a smart lighting algorithm. As revealed in our literature review, the existing work on intelligent lighting systems has primarily focused on hardware architecture, and the software developments are limited to on-off light settings with no capability to learn user preferences. As such, our work presents several elements of novelty. 
To our knowledge, it is the first to explore the application of FIS and RL for light management.
It uses the domain knowledge to develop an intelligent fuzzy controller, and at the same time, provides a mechanism for the user to interact with the algorithm, and thus create a learning behaviour.
We were not able to find similar work with respect to both FIS and RL in similar applications.

It is worth noting that the results presented in this paper can be extended in several directions.
For example, an adaptive learning rate can be utilized to accelerate the learning process of the algorithm when the baseline FIS and user preferences are significantly different.
Mimicking human decisions is very complex and there are numerous factors that stimulate visual comfort.
While the baseline FIS developed here was based on thorough research in the literature and discussions with an industry partner active in developing aircraft lighting systems, a deeper study of human lighting preferences (e.g., light color, human eye conditions) can help design a more effective baseline FIS, subsequently reducing the gap between user preferences and the algorithm output, and reducing the need for prolonged learning.
Moreover, our user study included 18 participants.
While this relatively limited number of participants does not compromise the rigor of our developments, a larger user study can help improve our knowledge about humans' lighting preferences and potentially inform new design directions that can help accelerate the adoption of such algorithms in real practice.
 
In the future, we will work on integrating the developed algorithm with smart window systems and will conduct a larger user study.

\section*{Acknolwedgement}
This research was partially supported by the Natural Sciences and Engineering Research Council of Canada Discovery Grants program.
\bibliographystyle{elsarticle-num}
\bibliography{References}

\clearpage
\appendix
\section*{Supplementary Material}
The fuzzy rules for the rule-base are as follows.
\small
\begin{longtable}{lllll}
Age   & Activity              & DGI & Chronotype & Output \\
0-20  & Sleeping              & 14  & Morning       & D1    \\
0-20  & Sleeping              & 14  & Intermediate       & D1    \\
0-20  & Sleeping              & 14  & Evening       & D1    \\
0-20  & Sleeping              & 18  & Morning       & D1    \\
0-20  & Sleeping              & 18  & Intermediate       & D1    \\
0-20  & Sleeping              & 18  & Evening       & D1    \\
0-20  & Sleeping              & 22  & Morning       & D1    \\
0-20  & Sleeping              & 22  & Intermediate       & D1    \\
0-20  & Sleeping              & 22  & Evening       & D1    \\
0-20  & Sleeping              & 25  & Morning       & D5     \\
0-20  & Sleeping              & 25  & Intermediate       & D5     \\
0-20  & Sleeping              & 25  & Evening       & D5     \\
0-20  & Sleeping              & 29  & Morning       & D5     \\
0-20  & Sleeping              & 29  & Intermediate       & D5     \\
0-20  & Sleeping              & 29  & Evening       & D5     \\
0-20  & Eating                & 14  & Morning       & LU2    \\
0-20  & Eating                & 14  & Intermediate       & LU2    \\
0-20  & Eating                & 14  & Evening       & LU1    \\
0-20  & Eating                & 18  & Morning       & LU1    \\
0-20  & Eating                & 18  & Intermediate       & LU1    \\
0-20  & Eating                & 18  & Evening       & D1     \\
0-20  & Eating                & 22  & Morning       & LU1    \\
0-20  & Eating                & 22  & Intermediate       & D1     \\
0-20  & Eating                & 22  & Evening       & D1     \\
0-20  & Eating                & 25  & Morning       & D1     \\
0-20  & Eating                & 25  & Intermediate       & D2     \\
0-20  & Eating                & 25  & Evening       & D2     \\
0-20  & Eating                & 29  & Morning       & D2     \\
0-20  & Eating                & 29  & Intermediate       & D3     \\
0-20  & Eating                & 29  & Evening       & D3     \\
0-20  & Meeting/Entertainment & 14  & Morning       & LU3    \\
0-20  & Meeting/Entertainment & 14  & Intermediate       & LU3    \\
0-20  & Meeting/Entertainment & 14  & Evening       & LU2    \\
0-20  & Meeting/Entertainment & 18  & Morning       & LU2    \\
0-20  & Meeting/Entertainment & 18  & Intermediate       & LU2    \\
0-20  & Meeting/Entertainment & 18  & Evening       & LU1    \\
0-20  & Meeting/Entertainment & 22  & Morning       & LU2    \\
0-20  & Meeting/Entertainment & 22  & Intermediate       & LU1    \\
0-20  & Meeting/Entertainment & 22  & Evening       & LU1    \\
0-20  & Meeting/Entertainment & 25  & Morning       & LU1    \\
0-20  & Meeting/Entertainment & 25  & Intermediate       & D1     \\
0-20  & Meeting/Entertainment & 25  & Evening       & D1     \\
0-20  & Meeting/Entertainment & 29  & Morning       & D2     \\
0-20  & Meeting/Entertainment & 29  & Intermediate       & D3     \\
0-20  & Meeting/Entertainment & 29  & Evening       & D3     \\
20-40 & Sleeping              & 14  & Morning       & D1    \\
20-40 & Sleeping              & 14  & Intermediate       & D1    \\
20-40 & Sleeping              & 14  & Evening       & D1    \\
20-40 & Sleeping              & 18  & Morning       & D1    \\
20-40 & Sleeping              & 18  & Intermediate       & D1    \\
20-40 & Sleeping              & 18  & Evening       & D1    \\
20-40 & Sleeping              & 22  & Morning       & D1    \\
20-40 & Sleeping              & 22  & Intermediate       & D1    \\
20-40 & Sleeping              & 22  & Evening       & D1    \\
20-40 & Sleeping              & 25  & Morning       & D5     \\
20-40 & Sleeping              & 25  & Intermediate       & D5     \\
20-40 & Sleeping              & 25  & Evening       & D5     \\
20-40 & Sleeping              & 29  & Morning       & D5     \\
20-40 & Sleeping              & 29  & Intermediate       & D5     \\
20-40 & Sleeping              & 29  & Evening       & D5     \\
20-40 & Eating                & 14  & Morning       & LU3    \\
20-40 & Eating                & 14  & Intermediate       & LU2    \\
20-40 & Eating                & 14  & Evening       & LU2    \\
20-40 & Eating                & 18  & Morning       & LU2    \\
20-40 & Eating                & 18  & Intermediate       & LU2    \\
20-40 & Eating                & 18  & Evening       & LU1    \\
20-40 & Eating                & 22  & Morning       & LU1    \\
20-40 & Eating                & 22  & Intermediate       & LU1    \\
20-40 & Eating                & 22  & Evening       & LU1    \\
20-40 & Eating                & 25  & Morning       & LU1    \\
20-40 & Eating                & 25  & Intermediate       & D1     \\
20-40 & Eating                & 25  & Evening       & D1     \\
20-40 & Eating                & 29  & Morning       & D2     \\
20-40 & Eating                & 29  & Intermediate       & D2     \\
20-40 & Eating                & 29  & Evening       & D3     \\
20-40 & Meeting/Entertainment & 14  & Morning       & LU4    \\
20-40 & Meeting/Entertainment & 14  & Intermediate       & LU3    \\
20-40 & Meeting/Entertainment & 14  & Evening       & LU3    \\
20-40 & Meeting/Entertainment & 18  & Morning       & LU3    \\
20-40 & Meeting/Entertainment & 18  & Intermediate       & LU3    \\
20-40 & Meeting/Entertainment & 18  & Evening       & LU2    \\
20-40 & Meeting/Entertainment & 22  & Morning       & LU2    \\
20-40 & Meeting/Entertainment & 22  & Intermediate       & LU2    \\
20-40 & Meeting/Entertainment & 22  & Evening       & LU1    \\
20-40 & Meeting/Entertainment & 25  & Morning       & LU1    \\
20-40 & Meeting/Entertainment & 25  & Intermediate       & D1     \\
20-40 & Meeting/Entertainment & 25  & Evening       & D1     \\
20-40 & Meeting/Entertainment & 29  & Morning       & D2     \\
20-40 & Meeting/Entertainment & 29  & Intermediate       & D2     \\
20-40 & Meeting/Entertainment & 29  & Evening       & D3     \\
40-60 & Sleeping              & 14  & Morning       & D1    \\
40-60 & Sleeping              & 14  & Intermediate       & D1    \\
40-60 & Sleeping              & 14  & Evening       & D1    \\
40-60 & Sleeping              & 18  & Morning       & D1    \\
40-60 & Sleeping              & 18  & Intermediate       & D1    \\
40-60 & Sleeping              & 18  & Evening       & D1    \\
40-60 & Sleeping              & 22  & Morning       & D1    \\
40-60 & Sleeping              & 22  & Intermediate       & D1    \\
40-60 & Sleeping              & 22  & Evening       & D1    \\
40-60 & Sleeping              & 25  & Morning       & D5     \\
40-60 & Sleeping              & 25  & Intermediate       & D5     \\
40-60 & Sleeping              & 25  & Evening       & D5     \\
40-60 & Sleeping              & 29  & Morning       & D5     \\
40-60 & Sleeping              & 29  & Intermediate       & D5     \\
40-60 & Sleeping              & 29  & Evening       & D5     \\
40-60 & Eating                & 14  & Morning       & LU3    \\
40-60 & Eating                & 14  & Intermediate       & LU2    \\
40-60 & Eating                & 14  & Evening       & LU2    \\
40-60 & Eating                & 18  & Morning       & LU3    \\
40-60 & Eating                & 18  & Intermediate       & LU3    \\
40-60 & Eating                & 18  & Evening       & LU2    \\
40-60 & Eating                & 22  & Morning       & LU2    \\
40-60 & Eating                & 22  & Intermediate       & LU1    \\
40-60 & Eating                & 22  & Evening       & LU1    \\
40-60 & Eating                & 25  & Morning       & LU1    \\
40-60 & Eating                & 25  & Intermediate       & D1     \\
40-60 & Eating                & 25  & Evening       & D1     \\
40-60 & Eating                & 29  & Morning       & D1     \\
40-60 & Eating                & 29  & Intermediate       & D2     \\
40-60 & Eating                & 29  & Evening       & D3     \\
40-60 & Meeting/Entertainment & 14  & Morning       & LU3    \\
40-60 & Meeting/Entertainment & 14  & Intermediate       & LU2    \\
40-60 & Meeting/Entertainment & 14  & Evening       & LU2    \\
40-60 & Meeting/Entertainment & 18  & Morning       & LU3    \\
40-60 & Meeting/Entertainment & 18  & Intermediate       & LU3    \\
40-60 & Meeting/Entertainment & 18  & Evening       & LU2    \\
40-60 & Meeting/Entertainment & 22  & Morning       & LU3    \\
40-60 & Meeting/Entertainment & 22  & Intermediate       & LU3    \\
40-60 & Meeting/Entertainment & 22  & Evening       & LU2    \\
40-60 & Meeting/Entertainment & 25  & Morning       & LU1    \\
40-60 & Meeting/Entertainment & 25  & Intermediate       & D1     \\
40-60 & Meeting/Entertainment & 25  & Evening       & D1     \\
40-60 & Meeting/Entertainment & 29  & Morning       & D1     \\
40-60 & Meeting/Entertainment & 29  & Intermediate       & D2     \\
40-60 & Meeting/Entertainment & 29  & Evening       & D3     \\
60+   & Sleeping              & 14  & Morning       & D1    \\
60+   & Sleeping              & 14  & Intermediate       & D1    \\
60+   & Sleeping              & 14  & Evening       & D1    \\
60+   & Sleeping              & 18  & Morning       & D1    \\
60+   & Sleeping              & 18  & Intermediate       & D1    \\
60+   & Sleeping              & 18  & Evening       & D1    \\
60+   & Sleeping              & 22  & Morning       & D1    \\
60+   & Sleeping              & 22  & Intermediate       & D1    \\
60+   & Sleeping              & 22  & Evening       & D1    \\
60+   & Sleeping              & 25  & Morning       & D5     \\
60+   & Sleeping              & 25  & Intermediate       & D5     \\
60+   & Sleeping              & 25  & Evening       & D5     \\
60+   & Sleeping              & 29  & Morning       & D5     \\
60+   & Sleeping              & 29  & Intermediate       & D5     \\
60+   & Sleeping              & 29  & Evening       & D5     \\
60+   & Eating                & 14  & Morning       & LU2    \\
60+   & Eating                & 14  & Intermediate       & LU2    \\
60+   & Eating                & 14  & Evening       & LU1    \\
60+   & Eating                & 18  & Morning       & LU2    \\
60+   & Eating                & 18  & Intermediate       & LU1    \\
60+   & Eating                & 18  & Evening       & LU1    \\
60+   & Eating                & 22  & Morning       & LU2    \\
60+   & Eating                & 22  & Intermediate       & LU2    \\
60+   & Eating                & 22  & Evening       & LU1    \\
60+   & Eating                & 25  & Morning       & LU1    \\
60+   & Eating                & 25  & Intermediate       & LU1    \\
60+   & Eating                & 25  & Evening       & D1     \\
60+   & Eating                & 29  & Morning       & D1     \\
60+   & Eating                & 29  & Intermediate       & D2     \\
60+   & Eating                & 29  & Evening       & D3     \\
60+   & Meeting/Entertainment & 14  & Morning       & LU3    \\
60+   & Meeting/Entertainment & 14  & Intermediate       & LU2    \\
60+   & Meeting/Entertainment & 14  & Evening       & LU2    \\
60+   & Meeting/Entertainment & 18  & Morning       & LU3    \\
60+   & Meeting/Entertainment & 18  & Intermediate       & LU2    \\
60+   & Meeting/Entertainment & 18  & Evening       & LU2    \\
60+   & Meeting/Entertainment & 22  & Morning       & LU3    \\
60+   & Meeting/Entertainment & 22  & Intermediate       & LU3    \\
60+   & Meeting/Entertainment & 22  & Evening       & LU2    \\
60+   & Meeting/Entertainment & 25  & Morning       & LU1    \\
60+   & Meeting/Entertainment & 25  & Intermediate       & LU1    \\
60+   & Meeting/Entertainment & 25  & Evening       & D1     \\
60+   & Meeting/Entertainment & 29  & Morning       & D1     \\
60+   & Meeting/Entertainment & 29  & Intermediate       & D2     \\
60+   & Meeting/Entertainment & 29  & Evening       & D3    
\end{longtable}

\end{document}